\documentclass{article}

\usepackage{PRIMEarxiv}

\usepackage[utf8]{inputenc} % allow utf-8 input
\usepackage[T1]{fontenc}    % use 8-bit T1 fonts
\usepackage{hyperref}       % hyperlinks
\usepackage{url}            % simple URL typesetting
\usepackage{booktabs}       % professional-quality tables
\usepackage{amsfonts}       % blackboard math symbols
\usepackage{nicefrac}       % compact symbols for 1/2, etc.
\usepackage{microtype}      % microtypography
\usepackage{lipsum}
\usepackage{fancyhdr}       % header
\usepackage{graphicx}       % graphics
\usepackage[utf8]{inputenc}
\usepackage{multicol}
\usepackage{xcolor}
\usepackage{subfig}
\usepackage[utf8]{inputenc}
\usepackage{graphicx}
\usepackage{titlesec}
\usepackage{eso-pic}
\usepackage[framemethod=tikz]{mdframed}

\usepackage{geometry}
\usepackage{amsmath}
\usepackage{parskip}
\usepackage[official]{eurosym}
\usepackage{todonotes}
\usepackage{csquotes}

\usepackage{rotating}
\usepackage{lmodern}
\usepackage{setspace}

\graphicspath{{media/}}     % organize your images and other figures under media/ folder

\title{Mining Tweets to Predict Future Bitcoin Price}

\author{
  Ashutosh ~Hathidara\thanks{The work was done as part of Masters degree at Indiana University.}\\
  SAP AI \\
  \texttt{ashutosh.hathidara@sap.com} \\
   \And
  Gaurav Atavale \\
  Indiana University \\
  \texttt{gatavale@iu.edu} \\
  \And
  Suyash Chaudhary \\
  Indiana University \\
  \texttt{suschaud@iu.edu} \\
}

\begin{document}
\maketitle

\begin{abstract}
Bitcoin has increased investment interests in people during the last decade. We have seen an increase in the number of posts on social media platforms about cryptocurrency, especially Bitcoin. This project focuses on analyzing user tweet data in combination with Bitcoin price data to see the relevance between price fluctuations and the conversation between millions of people on Twitter. This study also exploits this relationship between user tweets and bitcoin prices to predict the future bitcoin price. We are utilizing novel techniques and methods to analyze the data and make price predictions.
\end{abstract}

% keywords can be removed

\section{Introduction}
\label{chap:introduction}
Cryptocurrency has been a constant source of attention for people in the last decade. Especially bitcoin, one of the cryptocurrencies which have increased investment interest significantly. Social media platforms like Twitter were flooded with tweets related to bitcoin and other cryptocurrencies. People have become more focused on identifying investment opportunities in crypto firms. But due to this, crypto markets have also become too volatile and fluctuating. The price for specific cryptocurrencies changes when a renowned celebrity or politician tweets about it. Positive or negative fluctuations depend upon the sentiments of tweets and the market conditions. Market conditions still affect very less as compared to people’s sentiments.  

In this project, we are analyzing tweets related to Bitcoin extracted from the Twitter feed. We are using the Bitcoin Tweets dataset \cite{bitcoin-dataset} on Kaggle. The dataset contains 16M tweets made by people around the world. Due to this, the dataset contains tweets from multiple languages and regions. The dataset has the most number of tweets in duration from 2016-01-01 to 2019-03-29. Although the dataset also contains tweets before and after this time interval, the data for that is very sparse. Because of this, we are only analyzing the tweets in the time duration mentioned above.

\begin{figure}[!htp]
    \centering
    \includegraphics[width=1.0\textwidth]{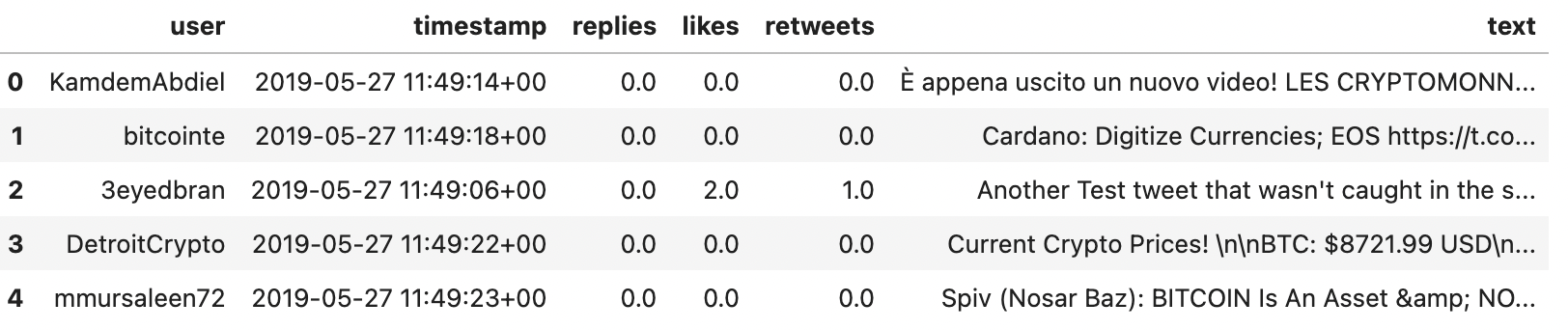}
    \caption{Raw data extracted from Kaggle}
    \label{fig:dataraw}
\end{figure}

Figure 1 illustrates the features in the dataset. Each row in the dataset consists of a single tweet. The metadata of the tweet is also given. Metadata consists of the username of the user who made the tweet, the timestamp when the tweet was made, and the traffic (replies, likes, retweets) on the tweet after it was made. The usernames provided are not fake or artificial. These are the real usernames of people at the time of data extraction. 

This report is structured in a way that first we will illustrate the methods we used for data mining-related tasks in the next section. Later in the results section, we will reveal the results we observed from the methods we employed.

\section{Methods}
\label{chap:methods}
Beginning with the raw dataset, we employed various preprocessing and analysis techniques. Later, we also utilized prediction techniques to validate whether the analysis we performed as part of the data mining tasks is actually helpful in real-time predictions. The below subsections explain the project setup and each of the methods we used for processing, analyzing, and predicting the data.

\subsection{Exploratory Data Analysis}
\label{sec:methods}
As the first step of the data mining process, we applied some preprocessing and analysis techniques to the dataset. As we mentioned earlier, the dataset contains tweets from many languages. We first applied language detection on the tweets using the python package langdetect \cite{langdetect}. We observed that ~80\% of the tweets in the dataset are English-language tweets. For flexibility, we filtered only English-language tweets for further analysis.

For analyzing this filtered dataset, we aggregated various features on a day level. Features like replies, likes, and retweets can be aggregated by adding them after grouping them on the day level. For tweet text, we have performed sentiment analysis \cite{sent-twitter} \cite{mood-stocks} to find the number of positive, negative, and neutral tweets every day. More about sentiment analysis is described in the next subsection. We also compared the trend for each of these features' tweet volume, likes, replies, and retweets from 2016 to 2019. We also segregated tweet volume w.r.t. the number of likes ($>0$, $>10$, $>100$, $>1000$) and observed the trend for that. We performed analysis on tweet volume by aggregating it to the hour of the day and day of the week independently to see if there is any particular day or hour when people are tweeting more about bitcoin.

As part of this step, we also preprocessed the data and made it ready for prediction purposes. We aggregated the data at a day level as described above and included all the features like day-level tweet volume, number of tweets with likes ($>0$, $>10$, $>100$, $>1000$), number of tweets with retweets ($>0$, $>100$), number tweets with the particular sentiment (positive, negative, neutral), one hot encoding the day of the week, one hot encoding hour of the day, price of bitcoin on the previous day, etc.

\subsection{Sentiment Analysis}
\label{sec:methods}
To understand the mood or opinion of the tweet, we performed sentiment analysis \cite{crypto-pred-sent} on our dataset. Tweets are generally noisy due to the inclusion of mentions, URLs, and emoticons. The tweet-preprocessor \cite{tweet-preprocessor} package was used to clean the data before sentiment analysis in order to counteract the influence of this noise in tweets. For sentiment analysis, we used two separate libraries: vaderSentiment \cite{vader-sentiment} and textblob \cite{textblob}. When compared to vaderSentiment, textblob labeled tweets more 'accurately'. We performed the sentiment analysis on the entire 'English' Bitcoin tweets for the selected duration.

\subsection{Clustering Analysis}
\label{sec:methods}
Before we actually went ahead with clustering, we performed quite a bit of preprocessing on our data, we started with dropping the text feature from the dataset as it was irrelevant to the clustering task followed by removing data that dated back to beyond 2016 and finally implementing another variation of clustering on the data i.e. we aggregated all the data into user level which gave us the total count of data features such as likes, comments, and retweets.

After this preprocessing, we experimented with three different approaches to Clustering namely K-means clustering \cite{kmeans}, Hierarchical Clustering \cite{hierarchical-clustering} and DBSCAN Clustering \cite{dbscan}. The data worked for K-means clustering without a hitch but for Hierarchical clustering and DBSCAN Clustering, we had to work with a fraction of our original data as the algorithms were running out of allocated memory.

\subsection{Regression}
\label{sec:methods}
As part of this step, we want to predict the bitcoin price of the next day based on various features we have after preprocessing the data. We will also include current-day bitcoin prices as part of this step. We have almost 3 years of data for tweets and we have aggregated it to day level. So, we have ~1100 (1 for each day) examples in our processed dataset. We have split this dataset into training and testing datasets where the testing split is 10\% of the total dataset. The data is split such that all the training data contains consecutive 90\% of the days and the last 10\% of days goes into the testing dataset. For standardizing the dataset, we used StandardScalar from sklearn \cite{sklearn}.

We employed different types of models for regression experiments. We experimented with simple Linear regression \cite{linear-regression}, Ridge regression \cite{ridge-regression}, and Lasso regression \cite{lasso-regression} models as part of classical regression models. We also experimented with tree-based regression models like Decision-Tree regression \cite{dt-regression} and Random Forest regression \cite{rf-regression}. Moreover, we also employed an artificial neural network \cite{ann-regression} approach for training the dataset. Some of the models described above require parameter tuning. We have used the K-fold cross-validation technique in combination with GridSearch for parameter tuning. We performed a comparative analysis of these models considering a common metric of comparison.

\subsection{Classification}
\label{sec:methods}
To predict if the price of bitcoin will dip or rise compared to the previous day, we have used various classification frameworks like KNN \cite{knn}, Logistic Regression model \cite{logistic-regression}, Naive Bayes model \cite{naive-bayes}, Kernel SVM \cite{svm}, Decision Tree Classification model, Decision Tree Classification model, Random Forest Classification, XGBoost \cite{xgboost}, and Light GBM.

The objective was to determine the direction of movement of the BTC price. The data and features remain the same as our Regression model. We have added a new classification Target Variable which takes the values of 0 and 1 based on price going down or up respectively. Here as well we kept train to test ratio as 90:10. We have performed k-fold cross-validation and grid search for all the frameworks. Accuracy, Recall, Precision and F1 scores are the metrics of measurement we have used here. Finally, we have plotted a ROC curve to compare the results of each model.

\section{Results}
\label{chap:results}
We have described the data processing and analysis processes in detail in the previous section. From the basic analysis of the day-level aggregation of the dataset, we have understood that the daily tweet volume and other tweet-related metrics (likes, replies, retweets) have increased from 2016 to 2019. Also, we have seen some regular impulses in the metric statistics where the metric values shoot up the graph. The reason behind that might be extraordinary fluctuation in the bitcoin price or other unusual hype in the cryptocurrency market. When we analyzed the number of tweets of certain sentiments (positive, negative, and neutral), we found out that the trend for each of them has been almost constant throughout the time (2016-2019). Comparatively, people have made more neutral tweets than positive and negative tweets as shown in figure 2 (a). We also performed analysis by observing the number of tweets on a certain day of the week. As shown in figure 2 (b), we can see that people are tweeting more about bitcoin on Fridays than any other day in the week. Additionally, we performed analysis to observe similar statistics about the hour of the day. We can observe in figure 2 (c) that the plot has a higher number of tweets from 9 am to 5 pm. This is usually work time and people are tweeting more about bitcoin during working hours.

\begin{figure}[!htp]
  \centering
  \subfloat[Sentiment Trend]{\includegraphics[width=0.7\textwidth]{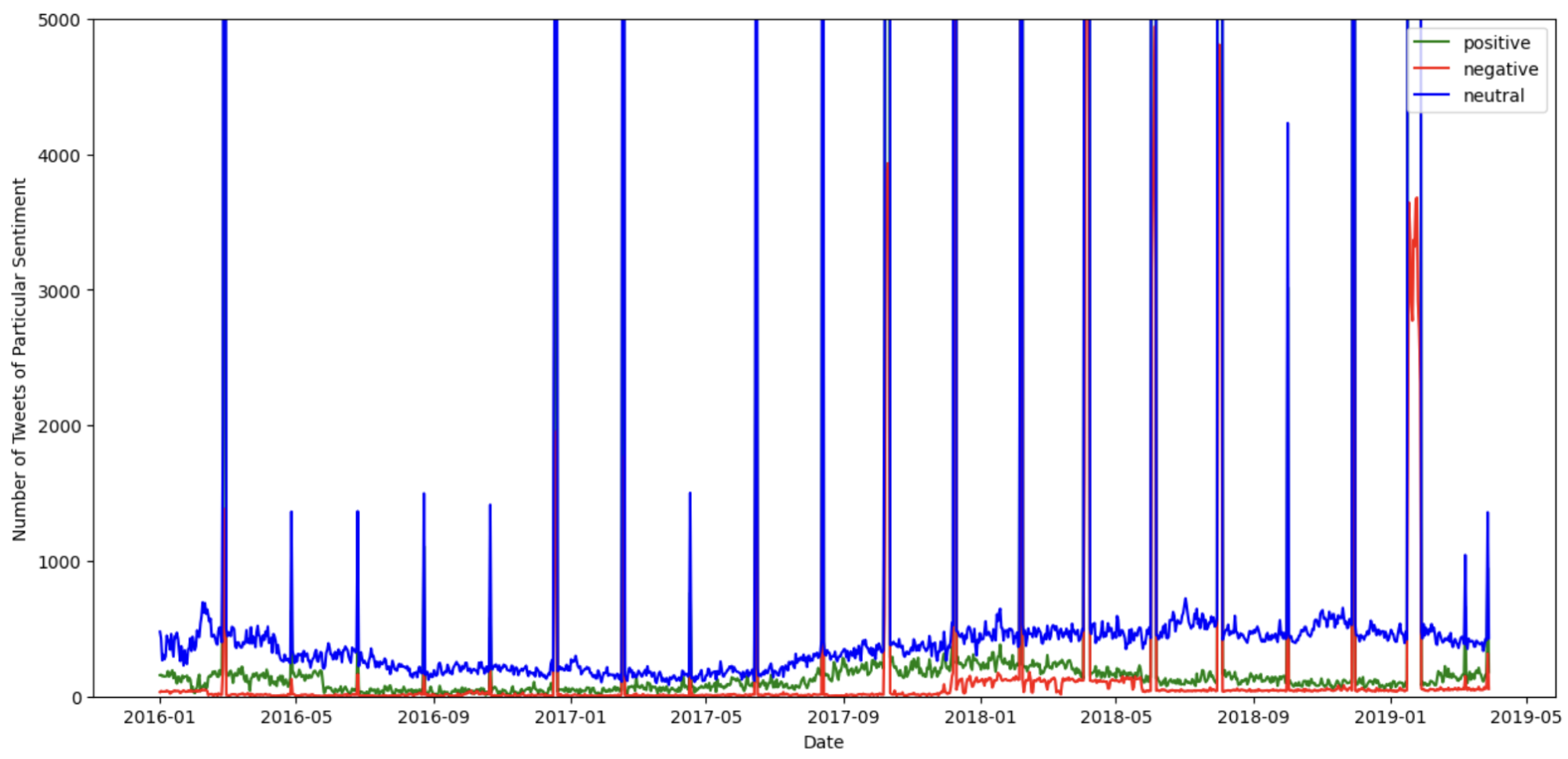}\label{fig:sentiment_eda}}
  \hfill
  \subfloat[Day-level Tweet Volume]{\includegraphics[width=0.7\textwidth]{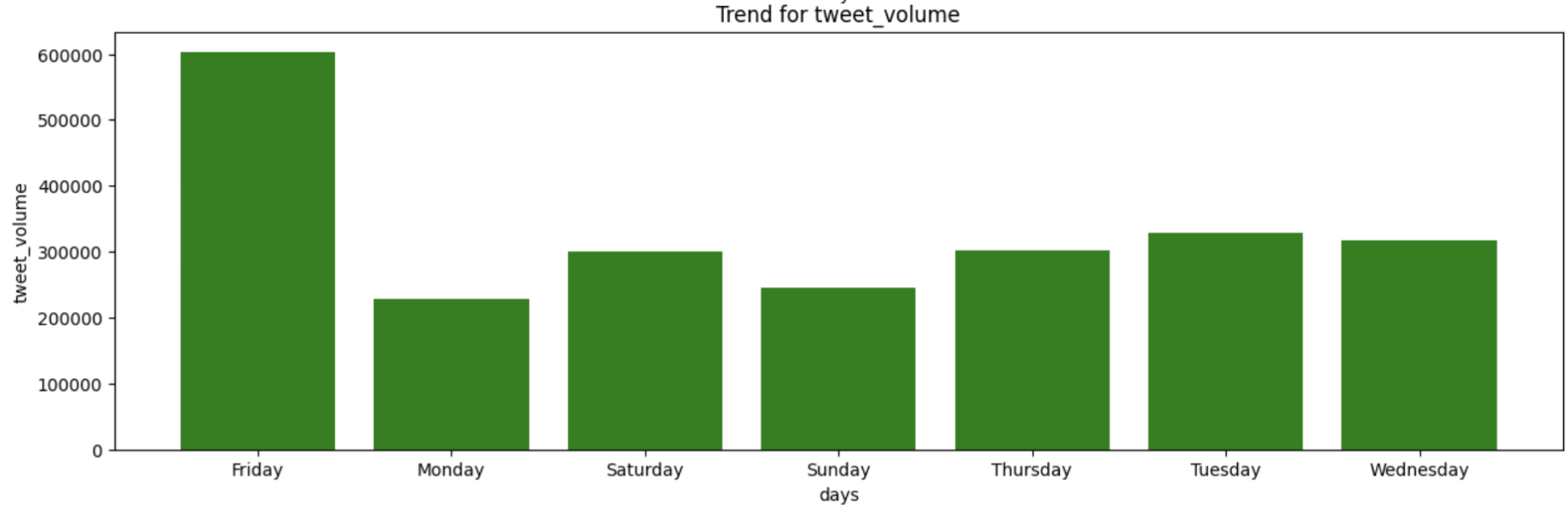}\label{fig:day_tweets}}
  \hfill
  \subfloat[Hour-level Tweet Volume]{\includegraphics[width=0.7\textwidth]{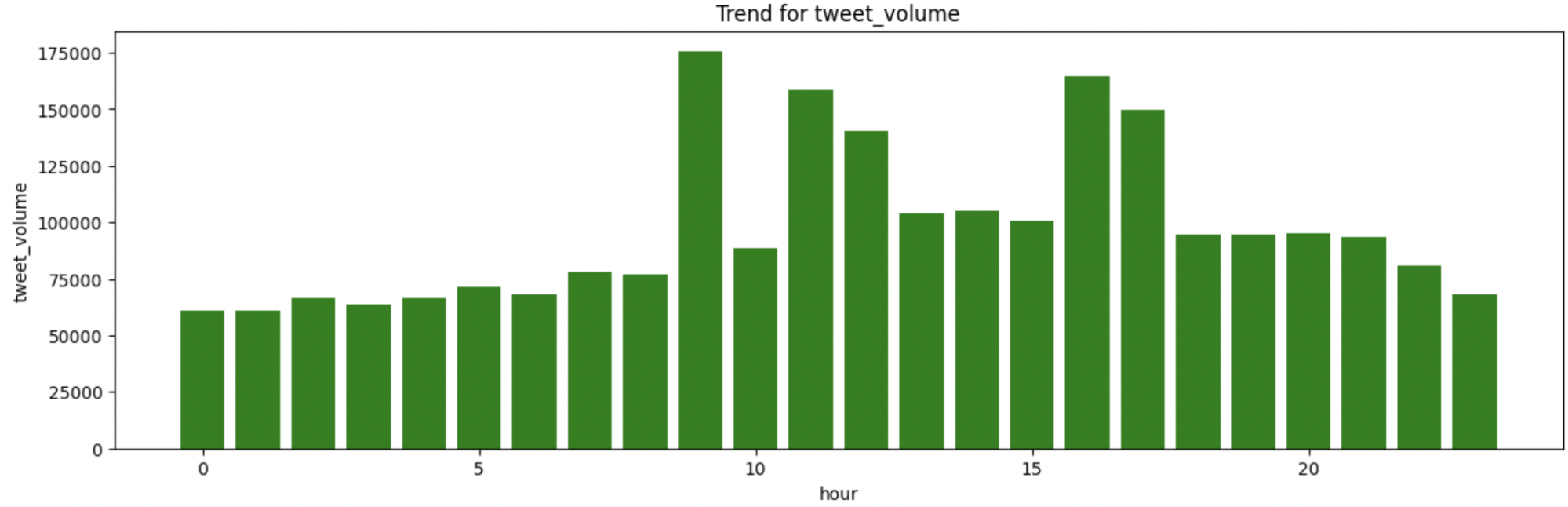}\label{fig:hour_tweets}}
  \caption{Results extracted from EDA}
\end{figure}

With respect to Sentiment Analysis, we could see that majority of the tweets are unbiased and informational. They don't share a specific emotion and are not biased towards positive or negative. More than 90\% of tweets fall into this category. Nearly 7\% of tweets are positive and ~3\% tweets are negative(distribution is shown in the figure below figure 3 (a)). We didn't find any correlation between sentiment and movement of price. To validate that the categorization was done right, we printed out a word cloud and we could visualize positive words dominating in the positive sentiment category and similarly negative words in the negative sentiment category. We see unbiased words in the Neutral category as shown in figure 3 (b).

\begin{figure}[!htp]
  \centering
  \subfloat[Distribution of Sentiments]{\includegraphics[width=0.45\textwidth]{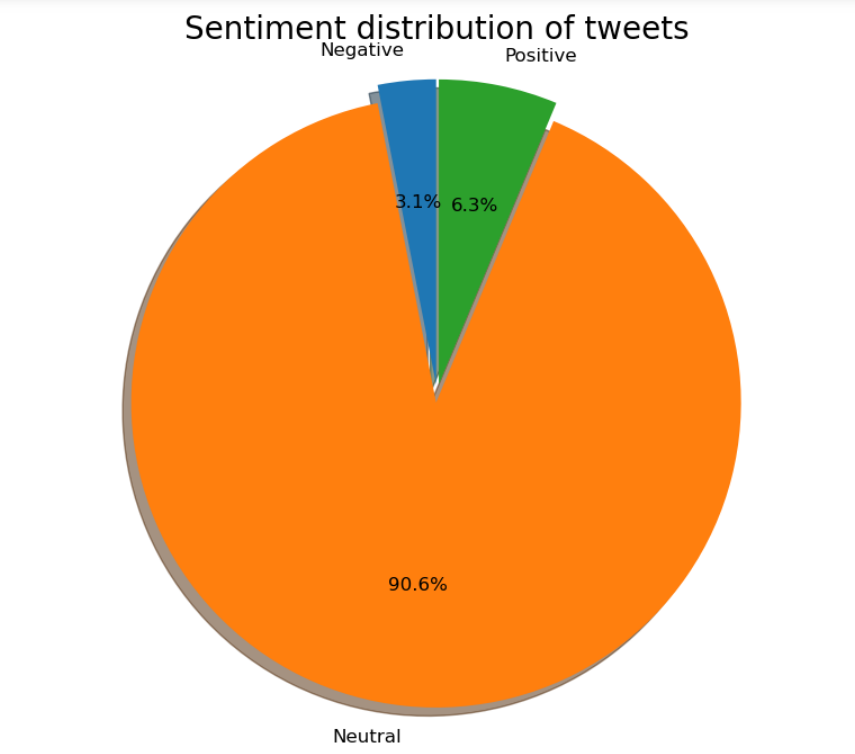}\label{fig:sent_pie}}
  \hfill
  \subfloat[Correlation with Bitcoin price]{\includegraphics[width=0.45\textwidth]{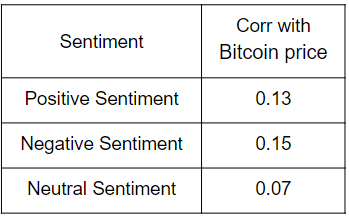}\label{fig:sent_corr}}
  \caption{Results extracted from Sentiment Analysis}
\end{figure}

If we talk about clustering, K-means Clustering yielded different results compared to what Hierarchical and DBSCAN Clustering produced, using evaluation metrics like distortion and inertia, we were able to plot the elbow curve and have k=3 as evident as shown in figure 4 (a) and (b) using K-means Clustering but for Hierarchical clustering and DBSCAN we received k=1 as the ideal k value based on the dendrogram we got for hierarchical clustering and the epsilon value we got from the elbow curve for DBSCAN. Hierarchical and DSCAN clustering might have produced different results because they weren’t really working with the entirety of data, the results might have been different if the algorithms supported computing the entire data but then again when you take the the original tweet text out of context, the remaining data might not produce desirable results when it comes to grouping users into separate groups. 

\begin{figure}[!htp]
  \centering
  \subfloat[Elbow curve for inertia]{\includegraphics[width=0.45\textwidth]{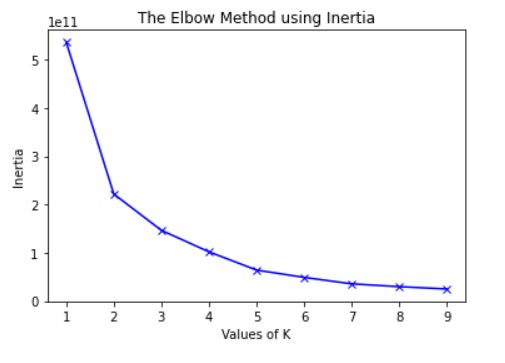}\label{fig:elbow1}}
  \hfill
  \subfloat[Elbow curve for distortion]{\includegraphics[width=0.45\textwidth]{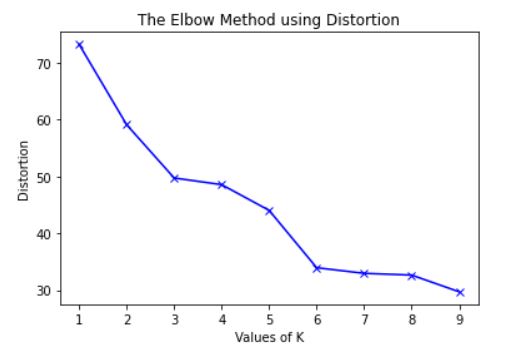}\label{fig:elbow2}}
  \caption{K-Means clustering analysis}
\end{figure}

As part of regression analysis, we performed the next-day bitcoin price prediction. We have used different types of models as described in the previous section. Using the simple regression models, we have found out that all 3 models linear regression, ridge regression, and lasso regression perform really well in predicting bitcoin price. Figure 5 (a) illustrates the results of ridge regression. The yellow scatter plot in the figure describes the original price data. The green line is the predicted price for training data and the red line is the predicted price for testing data. We are getting very similar results in the case of linear regression and lasso regression. In the case of tree-based models, we tried decision tree regression and random forest regression. Decision tree regression performs really well on the training dataset but performs very poorly on the test dataset. Random forest performs relatively better at testing datasets as compared to decision tree regression but both of these tree-based models perform worse than regression models. The reason behind the poor performance of tree based models can be because of splitting datasets and creating hard rules. These rules will not be generalizable for testing dataset. We also performed experiments on Artificial Neural Networks (ANN). We found out that it performs better than decision trees for the test dataset but it still performs poorly as compared to regression models. Figure 5 (b) illustrates the comparison of Mean Squared Error (MSE) on the test dataset of each of the models. We can see that the decision tree regression model performs the worst and the ridge regression performs the best.

\begin{figure}[!htp]
  \centering
  \subfloat[Actual BTC price \& Ridge regression predicted price]{\includegraphics[width=0.45\textwidth]{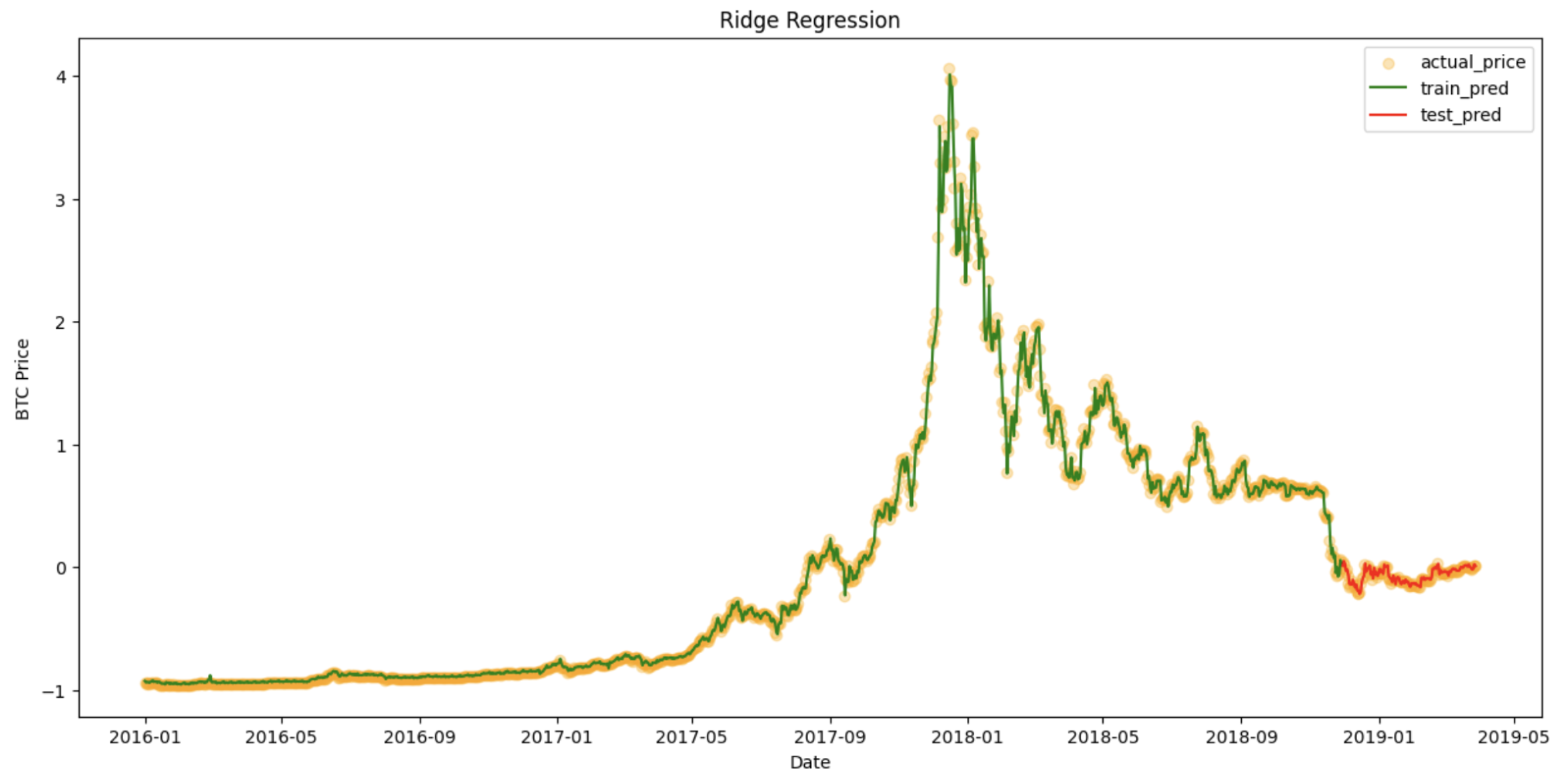}\label{fig:ridge_pred}}
  \hfill
  \subfloat[Comparison of MSE between different regression models]{\includegraphics[width=0.45\textwidth]{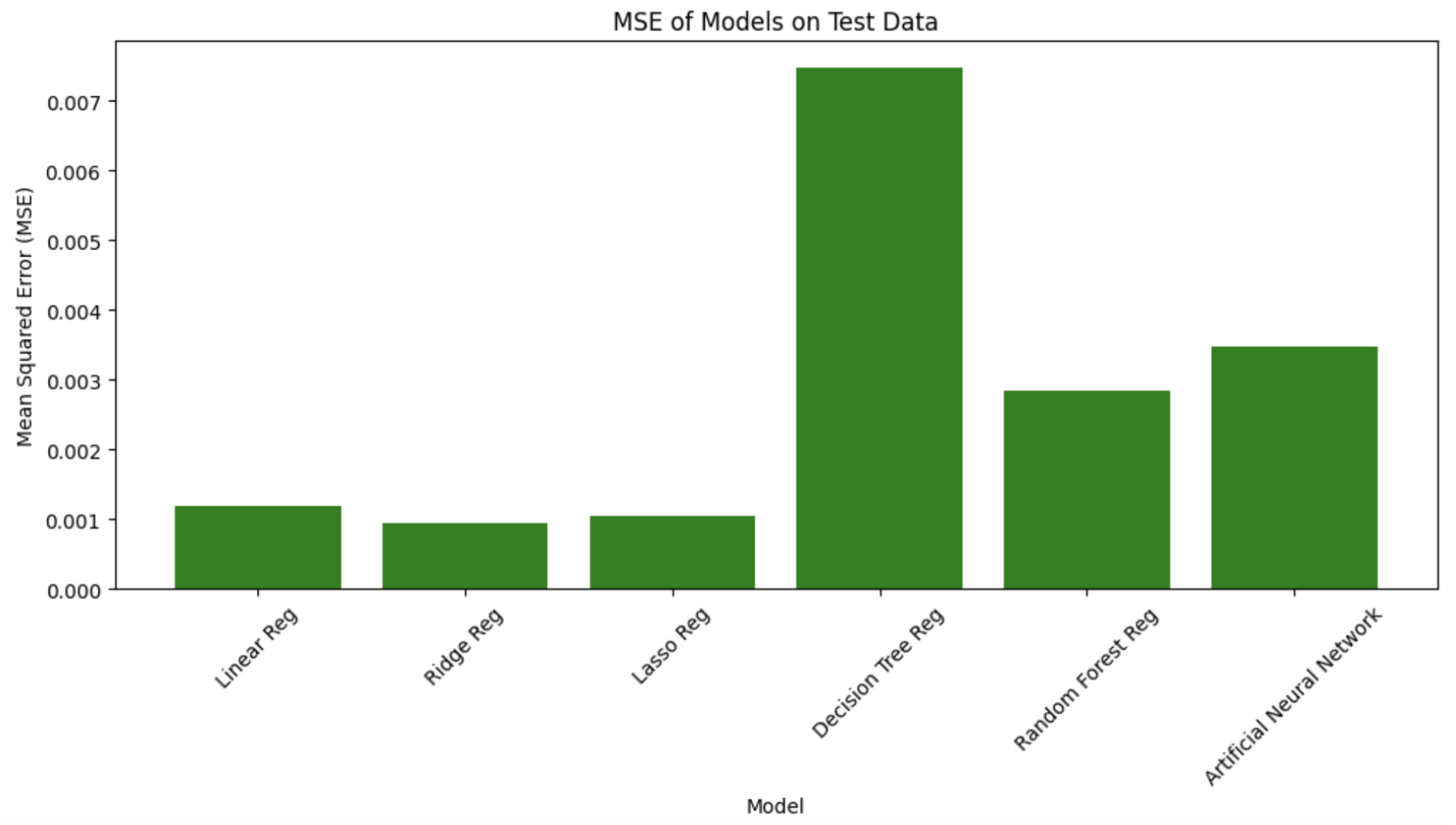}\label{fig:regression_comp}}
  \caption{Results extracted from Regression Analysis}
\end{figure}

\begin{figure}[!htp]
  \centering
  \includegraphics[width=0.7\textwidth]{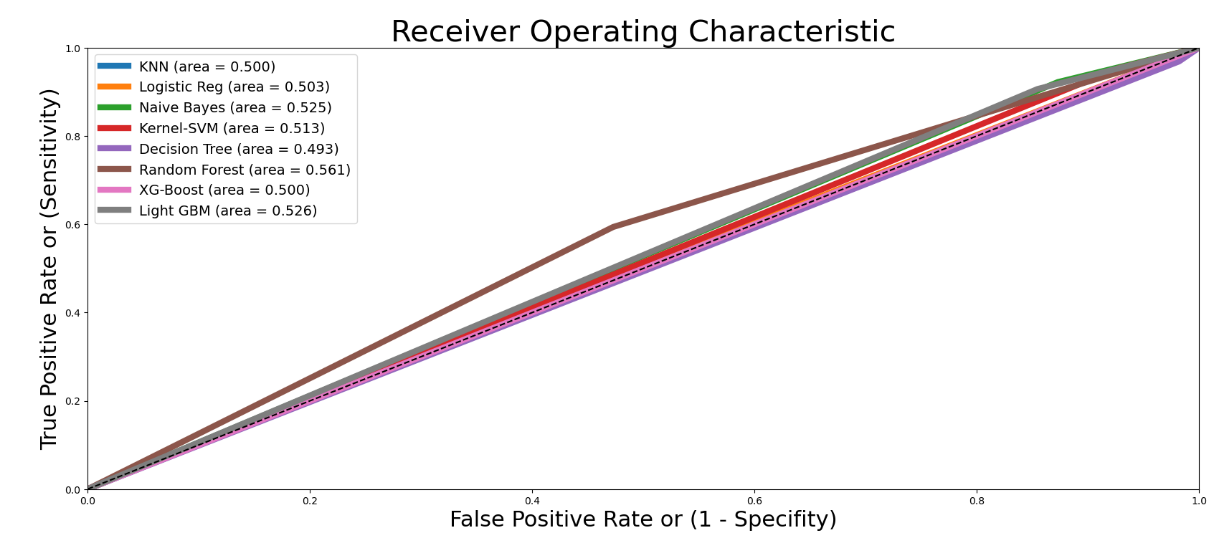}\label{fig:classification_comp}
  \caption{Comparing ROC curve of different classification models}
\end{figure}

Talking about classification analysis, We achieved the best results from Random forest classifier, which gave us accuracy of 62\% and F1 score of 75\%. The parameters with which we achieved these numbers are class\_weight:balanced, criterion:gini, max\_features:log2, n\_estimators: 100 and min\_samples\_split:2. The table 1 shows the scores for each model and figure 6 shows the Receiver operating characteristic (ROC) curve.

\begin{table}[!h]
\begin{center}
\begin{tabular}{|c|c|c|c|c|}
\hline
                    & \textbf{Accuracy} & \textbf{Precision} & \textbf{Recall} & \textbf{F1-score} \\ \hline
Logistic Regression & 0.49              & 0.54               & 0.41            & 0.46              \\ \hline
KNN                 & 0.54              & 0.54               & 1               & 0.7               \\ \hline
Naive Bayes         & 0.56              & 0.55               & 0.92            & 0.69              \\ \hline
Kernel SVM          & 0.55              & 0.54               & 0.95            & 0.69              \\ \hline
Decision Tree       & 0.53              & 0.53               & 0.97            & 0.69              \\ \hline
Random Forest       & 0.62              & 0.64               & 0.86            & 0.75              \\ \hline
XGBoost             & 0.54              & 0.54               & 1               & 0.7               \\ \hline
Light GBM           & 0.55              & 0.55               & 0.91            & 0.69              \\ \hline
\end{tabular}
\caption{\label{demo-table}Scores for models involved in Classification Analysis}
\end{center}
\end{table}

\section{Conclusion \& Discussion}
\label{chap:conc_discuss}
As we can see from the data, the BTC price field shows extremely high variation for this timeframe. Lowest value is \$364 and highest is \$19497 exhibiting a range of \$19K. This variation in data makes it hard to predict the exact price of BTC the next day through regression techniques. Also, as the fluctuations are huge and frequent, this poses a serious problem to classify the movement to see if price is going up or down. After an extensive EDA and understanding the data, we tried several techniques for regression and classification to achieve these objectives and have achieved decent results. We could further improve them by collecting more data and expanding the feature set. Some of the features which would help us could be- verified flag, google trends data, authentic news channels/accounts flag etc.

We could also deploy a similar methodology to predict prices for other cryptocurrencies as well. Some of the cryptocurrencies have few variants of hashtags which has to be taken into consideration while collecting tweets data.

We observed through clustering that the users could potentially be grouped into three categories namely, users who positively affect the price, users who negatively affect the price and users who have no effect on the price whatsoever. But the results of Clustering are something to be skeptical about as each algorithm produced different results due to lack of computation ability on the clustering algorithms among other reasons.

Lastly, we want to make a strong argument with this study that the user related tweets affect the price fluctuations of cryptocurrencies like Bitcoin and we can predict the price variation and even exact price using some of the features extracted from day-level aggregation of tweets.

%Bibliography
\bibliographystyle{unsrt}  
\bibliography{references}

\end{document}